\documentclass{llncs}
\usepackage{makeidx}  % allows for indexgeneration
\usepackage{hyperref}
\usepackage[utf8]{inputenc}
\usepackage{graphicx}
\usepackage{csvsimple}
\usepackage{multirow}
\usepackage{multicol}
\usepackage{bm}
\usepackage[T1]{fontenc}
\usepackage{float}
\usepackage{verbatim}
\usepackage{placeins}
\usepackage{tikz}

\usepackage{textcomp}
\usepackage[font={scriptsize,it}]{caption}

\usepackage{algorithm}
\usepackage{algorithmicx}
\usepackage{algpseudocode}
\usepackage{wrapfig}
\usepackage{subcaption}

\usepackage{subcaption}
\usepackage{mwe}

\usepackage{amsmath}
\usepackage{graphicx}
\usepackage[colorinlistoftodos]{todonotes}
\newcommand{\specialcell}[2][c]{%
  \begin{tabular}[#1]{@{}c@{}}#2\end{tabular}}

\begin{document}
\frontmatter
\pagestyle{headings}
\mainmatter

%\title{Scalability of Large Scale Distributed Deep Reinforcement Learning: a Case Study in Training Atari Games Agents}
\title{Distributed Deep Reinforcement Learning: learn how to play Atari games in 21 minutes\thanks{This research was supported in part by PL-Grid Infrastructure, grant identifier \texttt{rl2algos}}}
\author{Igor Adamski \inst{1}\thanks{All authors contributed equally to this work.}, Robert Adamski\inst{2,4}, Tomasz Grel\inst{1}, Adam Jędrych\inst{1}, \mbox{Kamil Kaczmarek\inst{1}} and Henryk Michalewski\inst{1,3}}

\institute{deepsense.ai\and 
Intel \and
Polish Academy of Sciences \and
Biz On Sp. z o.o.
}

\maketitle

\begin{abstract}
We present a study in Distributed Deep Reinforcement Learning (DDRL) focused on scalability of a state-of-the-art Deep Reinforcement Learning algorithm known as Batch Asynchronous Advantage Actor-Critic (BA3C). We show that using the Adam optimization algorithm with a batch size of up to 2048 is a viable choice for carrying out large scale machine learning computations. This, combined with careful reexamination of the optimizer's hyperparameters, using synchronous training on the node level (while keeping the local, single node part of the algorithm asynchronous) and minimizing the model's memory footprint, allowed us to achieve linear scaling for up to 64 CPU nodes. This corresponds to a training time of 21 minutes on 768 CPU cores, as opposed to the 10 hours required when using a single node with 24 cores achieved by a baseline single-node implementation.
\end{abstract}


\section{Introduction}

Gradient descent optimization is an indispensable element of solving many real-world problems including but not limited to training deep neural networks \cite{Goodfellow_deep_learning,optimization_reference}. Because of its inherent sequentiality it is also particularly difficult to parallelize \cite{MLHPC_scalability_limits}. Recently a number of advances in developing distributed versions of gradient descent algorithms have been made \cite{fair_paper,google_brain_paper,LARS,100_epoch_alexnet,TernGrad}. However, most of them deal with relatively simple variants of the algorithm, for which using larger batch sizes and increasing the learning rate (step size) often yield satisfactory results. 

In the case of tasks encountered in Deep Reinforcement Learning these simple optimization procedures are often insufficient and so more advanced algorithms such as RMSProp \cite{rmsprop_paper} and Adam \cite{adam_paper} are used more often \cite{deepmind_nature,deepmind_a3c,ga3c_2017,ba3c_paper,ppo}. However, these have not yet been subject to extensive formal analysis or even tests in largely distributed settings. This is crucial since the usual tasks of training models for reinforcement learning are often extremely computationally expensive \cite{deepmind_nature}. Therefore distributed training is gaining more and more traction in the supervised learning community. Devising efficient ways of distributing advanced variants of SGD has the potential to speed up the progress of the entire field. 

As our benchmarking task we chose the Atari 2600 emulator \cite{ALE} provided by the OpenAI Gym framework \cite{openai_gym} and the wide variety of games it offers. Atari games are considered a viable benchmark for testing deep reinforcement learning algorithms \cite{ga3c_2017,ppo,ba3c_paper}. The early attempts to develop agents that would efficiently play Atari games were presented in \cite{dqn}. This algorithm required as much as 8 days of training on a GPU \cite{deepmind_a3c} to reach a level that surpassed a casual human player. Later developments of the Asynchronous Advantage Actor-Critic (A3C) algorithm \cite{deepmind_a3c} reduced the learning time to several hours. Because of the work presented in \cite{ba3c_paper} a version of this algorithm optimized for Intel\textsuperscript{\textregistered}\ CPUs was already publicly available. A brief discussion of the single-node version of the algorithm is presented in section \ref{subsection_ba3c_background}.

In this work we present a distributed version of this algorithm that achieves linear scaling for the tested games for configurations of up to 64 nodes (see figure \ref{fig_scaling}). This allowed us to reduce the training time from roughly 10 hours to around 20 minutes while preserving the original accuracy of the models obtained. For a comparison with other similar implementations we refer the reader to table \ref{ga3c_comparison}.

Our contribution applies and extends the recent advances \cite{fair_paper,google_brain_paper} in distributed supervised learning to the field of reinforcement learning. Sections \ref{subsection_architecture}, \ref{section_sync_async}, \ref{section_optimizers} and \ref{subsection_communication_overhead} report on the design choices we made and the results they yielded. 
We also make our source code available for anyone who would like to reproduce or improve upon our results. Detailed instructions about running the experiments are also provided\footnote{The source code along with game-play videos  can be found at: \url{https://github.com/deepsense-ai/Distributed-BA3C}.}.

\subsection{Related work}

While distributed machine learning has recently been a topic of extensive research, it has mainly focused on supervised learning. For an in-depth review of scalability of modern supervised learning approaches, we refer the reader to \cite{MLHPC_scalability_limits}. This work also lists common problems with various approaches to distributing various gradient descent optimization procedures. The pitfalls identified include the communication overhead arising from the necessity to share the weight updates between the nodes. The authors concluded that using larger batches and step sizes had the potential to solve this problem but resulted in less accurate models. 

\textit{Relation to \cite{google_brain_paper}.}
Research done in \cite{google_brain_paper} delves more into the architectural aspects of distributed learning, by proposing to abandon the asynchronous design in favor of a synchronous one. It also makes detailed arguments about the problem of ``stale gradients'', which prevents the asynchronous paradigm from scaling beyond several nodes. We set out to verify these claims by performing our reinforcement learning experiments using both synchronous and asynchronous training in section \ref{section_sync_async}. For a survey of the various asynchronous gradient descent procedures we refer the reader to \cite{MLHPC_asynchronous_sgd} and \cite{MLHPC_asynchronous_core}.

\textit{Relation to \cite{fair_paper}.} 
Work done in \cite{fair_paper} focuses on large-scale supervised learning. It showed that a setup with many machines working concurrently can effectively speed up the training by a large margin. As a result of parallelizing multiple GPUs and using appropriate learning rates for effectively larger batches, training Resnet-50 on Imagenet was completed in 1 hour. The work \cite{fair_paper} also showed that using very large batch sizes requires rethinking the optimization algorithms used. In \cite{fair_paper} authors focused on the SGD with momentum optimizer which often works very well in supervised learning  \cite{sutskever_momentum}. Our work attempts to apply similar principles to the Adam optimizer which we found more suitable for reinforcement learning tasks. The details can be found in section \ref{section_optimizers}.

The authors of \cite{LARS}  recognized the need to modify the optimization procedures in order to better utilize the distributed settings. The modification proposed a novel procedure called ``Layer-wise Adaptive Rate Scaling'', which enabled efficient training of supervised learning models with batch size of up to 32768. Deploying this algorithm to the task of training large convolutional nets in \cite{100_epoch_alexnet} yielded extremely competitive training times of 24 minutes, as opposed to 1 hour achieved without these modifications in \cite{fair_paper}.

An interesting approach to reducing the communication overhead by ternarizing the gradients was recently proposed in \cite{TernGrad}. This is a part of larger research aiming at gradient quantization i.e., reducing the precision of the communicated values \cite{quantization1} \cite{1BitSGD}). A related approach is gradient sparsification, i.e. refraining from the exchange of small gradients, see e.g., \cite{sparse_gradients1}, \cite{sparse_gradients2}). However, both  quantization and sparsification drastically change the flow of training of a neural model. Since Reinforcement Learning training is already quite complex, we refrained from employing these methods. Still, they certainly should be considered in future DDRL experiments. 

To date, only limited formal research has been done in optimizing and parallelizing targeted strictly at reinforcement learning procedures. Notable works in this domain include \cite{deepmind_gorilla}, which focused on reducing the long training times observed in \cite{deepmind_nature}. A significant speedup (by an order of magnitude \cite{deepmind_gorilla}) and higher game scores were achieved. This was done using large resources of up to 130 nodes, by applying the Asynchronous SGD paradigm to the model developed in \cite{deepmind_nature} in a manner similar to the work focusing on supervised learning presented in \cite{DistBelief}.

Further work in \cite{deepmind_a3c} applied the asynchronous paradigm to the policy optimization methods, resulting in the Asynchronous Advantage Actor-Critic algorithm (A3C). These experiments used the relatively low computing power of a 16-core CPU. The work in \cite{ba3c_paper} sought to optimize a more efficient batched variant of this algorithm for use with commodity Intel\textsuperscript{\textregistered}\ Xeon CPUs by employing the Math Kernel Library. A GPU-based version of this algorithm has also been presented in \cite{ga3c_2017}. None of these works explicitly dealt with communication overheads in distributed policy optimization.

Significant computing resources were used in the work presented in \cite{deepmind_alpha_go_nature_2016} to develop AlphaGo -- a program for playing the game of Go. This work was based on a combination of reinforcement learning, supervised learning and tree search methods. The authors reported using configurations of up to 1920 CPUs and 280 GPUs for testing the algorithm which provided a significant improvement in the quality of the results achieved. It is also mentioned that the training of the policy network was done using 50 GPUs for one day \cite{deepmind_alpha_go_nature_2016}.

Further work in \cite{deepmind_alpha_go_zero} focused on achieving better results without using supervised learning and handcrafted features. Computationally, the training utilized the synchronous paradigm with 64 GPU workers and 19 parameter servers, using a total batch size of 2048. For optimization the Momentum SGD optimizer with learning rate annealing was used. Recently this work has been further extended in \cite{deepmind_alpha_zero} where the authors presented a general algorithm able to achieve expert level also in chess and shogi. Notably the training was completed in 24 hours and a relatively large batch size of 4096 was used.

Recently a novel algorithm called Proximal Policy Optimization (PPO) was proposed in \cite{ppo}. Notably it also uses the Adam optimizer on which we focused in this work and achieved promising scores in Atari games. A distributed version of this algorithm was used in \cite{openai_sumo}, where the authors used 4 GPUs, Adam optimizer and batch size of 5120. Further examination of PPO in a distributed setup appears as a promising area of future research.

A different approach to distributing reinforcement learning has recently been presented in \cite{es_paper}. In this work the parallelization was applied to an evolution strategy (ES), which is a direct search method. This property enables efficient exchange of information between the workers since they only have to communicate scalar values. Because of that the algorithm is especially easy to distribute since the communication costs of sharing the gradient updates don't apply to this scheme. When training a 3D humanoid to walk the authors reported linear scaling for up to 1440 CPU cores \cite[p.~8]{es_paper}. After 1 hour of training agents for Atari games on 720 CPUs evolution strategies were able to achieve scores comparable to the ones achieved by A3C (which was trained for 24 hours with a single CPU) \cite[p.~7]{es_paper}.

Detailed analysis of Deep Reinforcement Learning on a single machine with multiple GPUs was recently published in \cite{accelerated_methods}. The authors reported using a batch size of up to 2048 and utilizing 8 GPUs for various Deep RL algorithms. Impressive training times (under 10 minutes) are also reported.
\section{Distributed BA3C implementation}\label{section_2}
For all our experiments we used the Batch Asynchronous Advantage Actor-Critic algorithm implemented in \cite{Tensorpack} and later modified in \cite{ba3c_paper}. A similar algorithm was also described in \cite{ga3c_2017}. We will not elaborate on its properties here but rather focus on explaining the details behind implementing it on multiple parallel machines. A good description of the single-node version of this algorithm can be found in \cite{ba3c_paper}. Here we are using multiple clustered CPUs and each of them individually performs the BA3C algorithm. The individual nodes maintain a shared copy of the model through the use of special nodes called parameter servers, which store the model weights. Table \ref{hyperparameters_table} presents the hyperparameters of the algorithms, which were specifically tuned to achieve lowest training times during our research.
Some of the hyperparameters are omitted and for those we assume the default values used in \cite{ba3c_paper}. Detailed description of the hyperparameters describing the Adam optimizer is presented in section \ref{section_optimizers}.

\vspace{-1cm}
\renewcommand{\arraystretch}{1.2}
\begin{center}
\begin{table}[h]
\caption{Hyperparameters of the distributed BA3C implementation}
\label{hyperparameters_table}
\begin{tabular}{ |c|c|l|}
  \hline
 {Symbol} & \specialcell{Default\\value}&Description\\
 \hline
$n$ & 64 & Number of CPU nodes used for the distributed training\\
$c$ & 12 & Number of cores in each worker CPU\\
$\eta$    & 0.001 &   Learning rate (step-size)\\
\texttt{bs}  & 32   & Number of data-points in each training batch (on each node)\\
\texttt{ps} & 4 & Number of nodes responsible for holding the model parameters\\
\texttt{n\_sim} & 10 & Number of Atari simulators used simultaneously on one worker\\
$\epsilon$  & $10^{-8}$   & Constant used for numerical stability in the Adam optimizer\\
$\beta_1$, $\beta_2$ & 0.8, 0.75 & Decay rates of the running averages used in Adam optimizer\\
 \hline
\end{tabular}
\end{table}
\end{center}
\vspace{-1.2cm}

\subsection{BA3C background} \label{subsection_ba3c_background}
The BA3C design on a single node focuses on parallel interactions of multiple agents with game environments, that produce experience data later aggregated into mini-batches used for training. We call an ``agent'' an instance of the model interacting with the outside environment, in our context the Atari 2600 emulator.

The interaction with an Atari game is attained via OpenAI Gym \cite{openai_gym} providing the model with input of RGB image pixels and enabling the agents to act upon any state of the game. On a single machine, \texttt{n\_sim} simulators of the Atari environment are running concurrently and an agent plays one game on each of them. Each consecutive \texttt{frame\_hist} frames from the game count as one state. For each state of the game a policy query is sent to the prediction thread, which then feeds the input image to the neural net, outputting a respective behavioral guideline. The action performed is then gathered together with the state it was acted on and the reward it received from that action. This tuple creates a data-point.
Then, \texttt{bs} data-points are assembled into a mini-batch, that is back-propagated through the neural net giving gradients, later used to perform gradient descent on a cost function.

\subsection{Architecture} \label{subsection_architecture} Let us suppose that we are using $n$ workers. Each of the $n$ workers possesses a copy of the BA3C algorithm, which will parallelize the training among its $c$ cores. As aforementioned, a tuple of state, action and reward constitutes a single data-point, and a mini-batch of \texttt{bs} of those data-points is then used to compute the gradients which are synchronously gathered from all the workers, averaged and applied to the weights through the Adam optimizer.
All the weights are located in \texttt{ps} different parameter servers, each of which stores 1/\texttt{ps} of the model's parameters (4 convolution layers and 3 fully connected layers). The parameter servers then send the updated parameters back to the workers, which then again play the games and the process goes on until a satisfactory model is achieved.% This is visualized in figure \ref{fig:architecture}.
\vspace{-1cm}
\begin{figure}[h]
\centering
\input{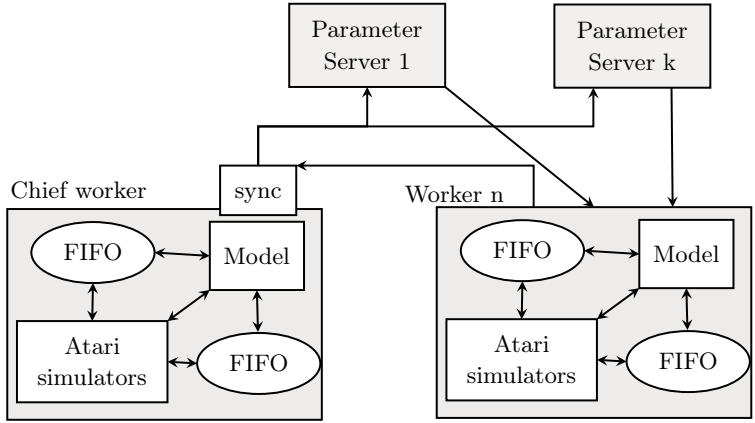}
\caption{Our approach to distributed learning. The figure shows the synchronous training architecture which was our final choice.}
\label{fig:architecture}
\end{figure}
\vspace{-1cm}

\subsection{Synchronous vs Asynchronous training} 
\label{section_sync_async}
\subsubsection{Background on gradient descent optimization.}
Training deep neural networks usually involves gradient descent optimization. This is convenient since the gradient of the model with respect to some chosen loss function can be easily obtained by backpropagation \cite{optimization_reference}. Gradient descent is an iterative algorithm that in each iteration attempts to modify the model's parameters $\bm\theta$ in order to achieve a lower value of the cost function $J(\bm\theta, \bm{x})$ for some training data $\bm{x}$. Given the gradient $\bm{g}_t$ of the cost function w.r.t. the model parameters $\bm\theta$ (which can be obtained from the backpropagation procedure) the basic update rule for obtaining the new values of parameters at time step $t$ given the old values $\bm\theta _{t-1}$ can be written as:
\begin{equation}
\bm{\theta} _{t} = \bm{\theta} _{t-1} - \lambda \bm{g}_t,
\end{equation}
where the $\lambda$ parameter controls the learning rate (step size).

Numerous improvements to this scheme have been proposed. For a broad overview of different approaches we refer the reader to \cite{optimization_reference}. Of the recent improvements the RMSProp \cite{rmsprop_paper} and Adam \cite{adam_paper} procedures are widely used in Reinforcement Learning \cite{deepmind_nature,deepmind_a3c,ga3c_2017,ba3c_paper,ppo}. In the course of our experiments we found that Adam performs better on our task and therefore we will not elaborate on RMSProp further. A brief description of Adam optimizer is given in section \ref{adam_background}.	

An important decision that largely influences the outcome of the model's performance is the way of parallelizing the work of multiple nodes. Data parallelism in gradient descent algorithms can be done in two ways: synchronously or asynchronously. We have found that when using a large number of distributed nodes, these two approaches produce completely different results.

\subsubsection{Asynchronous training.} 
In the asynchronous approach, each of the workers, after collecting a mini-batch of data points, computes gradients and then uses them to perform weight updates. The weights of the model reside in parameter servers, which receive gradients from the workers and send the updated copy of the current model to each training instance. Therefore each individual worker updates the commonly shared parameters of the model without delay as soon as it completes computing its gradient. This has several advantages, one is that compared to a single machine implementation our model is guaranteed to perform $k$ times as many updates, if we are using $k$ workers. Another is that because workers do not need to wait for others to finish but rather apply updates continuously, we are utilizing a lock-free paradigm that helps make the most of the processing power at our disposal.

\begin{figure}[H]
\centering
\input{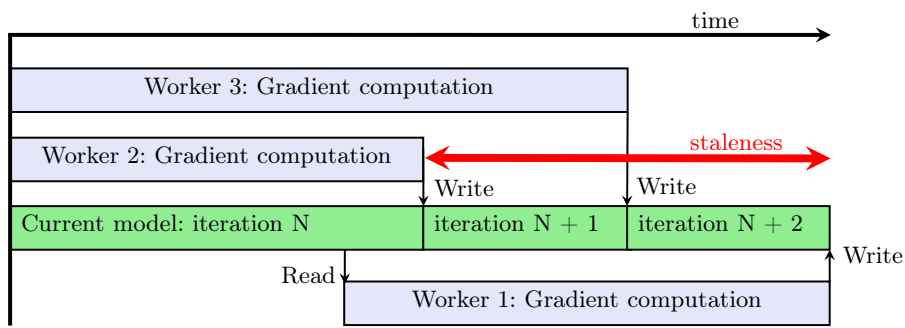}
\caption{A diagram representing gradient staleness - a systematic flaw related to asynchronous training with many workers.}
\label{fig:gradient_staleness}
\end{figure}

However, the pure asynchronous approach possesses also other characteristics that could impede the learning and prevent convergence. One such disadvantage is called stale gradients \cite{google_brain_paper}. It arises when a worker updates the weights using gradients that are outdated with regard to the current model. This is guaranteed to happen because during the time that the worker was processing the data and computing the gradients, the model has been updated several times by other nodes and now, when the worker applies the gradients, it will do so with respect to the model that is out-of-date. This is shown in the figure \ref{fig:gradient_staleness}.
\vspace{-0.5cm}

\begin{figure}
\begin{subfigure}[b]{0.50\textwidth}
\input{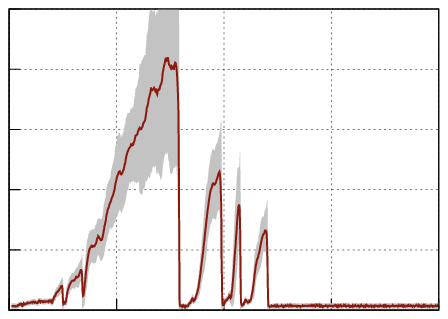}
\caption{Online score}
\label{async_instability_score}
\end{subfigure}
~
\begin{subfigure}[b]{0.50\textwidth}
\input{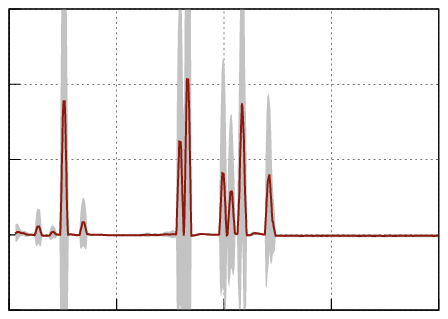}
\caption{Total training loss}
\label{async_instability_loss}
\end{subfigure}
\caption{Typical asynchronous training attempt, 64 workers.}
\end{figure}
\vspace{-1cm}

\subsubsection{Synchronous training.}
The synchronous training architecture is visualized in figure \ref{fig:architecture}. One of the workers (called a \textit{chief worker}) is special in a sense that it’s responsible for aggregating the gradients from all the others. A ``regular'' worker no longer posseses the power to update the model on its own -- it can only compute its gradient estimate and send it to the chief worker. Once enough gradients from the workers are accumulated, the chief worker updates the weights and sends the new them to the parameter servers. The new  weights are then sent to other workers and a new training iteration begins. This ensures that all the workers always have an up-to-date copy of the model weights, which solves the stale gradients problem.

The side effect of this procedure is the increase of the effective batch size used for performing a single update. Although we are not able to linearly increase the speed of model updates with the number of workers as in the asynchronous design, we expect the updates that are made to be more accurate since the gradient estimate is done using a larger batch size. This might in turn allow for larger step sizes to be used, which can hopefully compensate for the updates being less frequent and provide speed up. Importantly, synchronous training removes the problem of gradient staleness, as no worker computes gradients on an obsolete model, because updates are performed only after all of the workers compute their individual gradients.

The fact that we need to wait for all the workers can cause delays. This arises whenever, for various reasons, some of the workers may be lagging behind others in assembling their batches and computing gradients. We call this the \textit{slow stragglers problem}. Since the synchronous design imposes waiting for all the workers' gradients to perform a weight update, the effective time it takes for an update to occur is the time it takes the slowest worker to assemble his batch and compute the gradients. Therefore, reducing the number of gradients that we need to wait for to make an update could significantly reduce the influence of the slow stragglers. Detailed analysis of this phenomenon presented in \cite{google_brain_paper} confirms that waiting for around 90\% of gradients as opposed to all of them significantly improves the training times.

Another key fact that needs to be addressed when discussing synchronous training is the large effective batch size\footnote{We use the term \textit{effective batch size} to denote the number of training samples participating in a single weight update. In synchronous training this is equal to the local batch size on each node multiplied by the number of workers required to perform an update. In asynchronous training effective batch size is equal to the local batch size.} it tends to create. Since we are using $n$ mini-batches of data-points from every worker and then averaging them, we are virtually using a single batch of size $n\times \texttt{bs}$ to perform a single update. This may indicate the need to adjust other hyperparameters of the algorithm such as the learning rate. We revisit this issue in section \ref{section_optimizers}.

With the slow stragglers problem removed, the synchronous approach is much more intuitive and reasonable -- it does not risk gradient staleness and the weight updates that are made are much more accurate and less noisy. Making just one update for all the workers working on a model assures that the nodes are working collectively and efficiently upon a goal, whereas the asynchronous learning strategy seems rather chaotic and unstructured.

We performed a series of experiments to determine which paradigm is better in our use case. We found that asynchronous training causes large instabilities in the learning process. One of such experiments is shown in the figures \ref{async_instability_score} and \ref{async_instability_loss}. In this experiment the learning was proceeding correctly until after about 50 minutes the online score\footnote{By \textit{online score} we refer to the scores obtained by the agent during training. By contrast an \textit{evaluation score} would be a score obtained during the test phase. These scores can differ substantially, because while training the actions are sampled from the distribution returned by the policy network (this ensures more exploration). On the other hand, during test time the agent always chooses the action that gives the highest expected reward. This usually yields higher scores, but using it while training would prevent exploration.}
dropped suddenly to zero. This coincided with a large spike in the total training loss visible in the figure \ref{async_instability_loss}. We suspect it is caused by the stale gradients. We did not encounter this phenomenon when using synchronous training, therefore we have chosen to work with the synchronous architecture.

\subsection{Optimizer changes}\label{section_optimizers}

\subsubsection{Background on Adam optimizer.} \label{adam_background}
Adam optimizer was first described in \cite{adam_paper} and can be thought of as extension of the works presented in 
\cite{rmsprop_paper} and \cite{adagrad}. It maintains the exponentially decaying running averages $m _t$ and $v _t$ of all the previous gradients and squared gradients:
\begin{align}
%\bm{m}_t&=(1-\beta_1)\sum_{i=1}^t \beta_1^{t-i}\bm{g}_i \\
{m}_t&=\beta _1 m _{t - 1} +  (1-\beta_1) g _t,  \\
%\bm{v}_t&=(1-\beta_2)\sum_{i=1}^t \beta_2^{t-i}\bm{g}_i^2
{v}_t&=\beta _2 v _{t - 1} +  (1-\beta_1) g _t ^2
\end{align}
It then perform bias correction to define ${\hat{m}}_t={m}_t/(1-\beta_1^t)$ and ${\hat{v}}_t={v}_t/(1-\beta_2^t)$ and gives the final weight update for parameter $\theta$ at timestep $t$ as:
\begin{align} \label{adam_update_rule}
{\theta}_{t}={\theta}_{t-1} -\eta\dfrac{{\hat{m}}_t}{\sqrt{{\hat{v}}_t}+\epsilon}
\end{align}
Instead, many implementations (including TensorFlow \cite{tensorflow2015-whitepaper}) use less clear but more efficient formulation:
\begin{align}
\eta _t &= \eta \dfrac{\sqrt{1 - \beta _2 ^t}} {1 - \beta _1 ^t}, \\
\theta _t &= \theta _{t - 1} - \eta _t \dfrac{m _t} {\sqrt{v_t} + \hat{\epsilon}} \label{adam_update_rule_optimized}
\end{align}

The $\hat{\epsilon}$ in the equation \ref{adam_update_rule_optimized} and $\epsilon$ in equation \ref{adam_update_rule} are added for numerical stability, not to divide by 0 in the first timestep.

This means that the algorithm in this formulation has 4 hyperparameters that need tuning: the learning rate $\eta$, the decay factors for the running averages: $\beta _1$ and $\beta _2$ and $\hat\epsilon$. Next section provides insight into how these might need to be modified when transitioning from a single-node to a multi-node configuration.

\subsubsection{Increasing the learning rate.} Using very large batches that result from utilizing a large number of workers in the synchronous paradigm poses some challenges on the selection of optimizer hyperparameters. This problem is especially severe when distributing an algorithm that already had its hyperparameters chosen carefully.

Research on large scale distributed SGD by \cite{fair_paper} has addressed this problem by deploying the linear scaling rule: when multiplying the mini-batch size by $k$, multiply the learning rate by $k$. However this was done using much simpler SGD with momentum optimizer. We on the other hand have experimented with multiple optimizers and have found that only Adam \cite{adam_paper} and occasionally RMSProp \cite{rmsprop_paper} have brought about positive results in the asynchronous design.

\begin{figure}
\begin{subfigure}[b]{0.50\textwidth}
\input{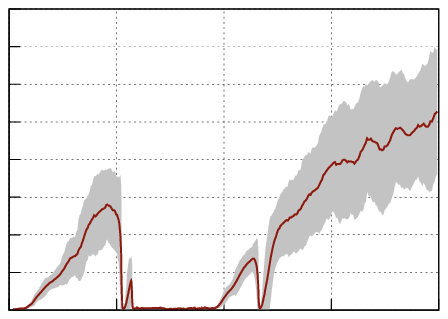}
\caption{Online score}
\label{lr_online_score}
\end{subfigure}
~
\begin{subfigure}[b]{0.50\textwidth}
\input{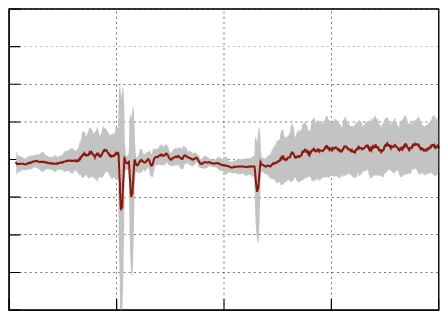}
\caption{Total training loss}
\label{lr_total_loss}
\end{subfigure}
\caption{Experiment with a learning rate $\eta = 0.002$. 64 nodes, synchronous training, local batch size of 64, total batch size of 4096. Increasing the learning rate for the Adam optimizer from 0.001 to 0.002 causes large instabilities clearly visible on the online score plot (figure \ref{lr_online_score}) and the total training loss plot (figure \ref{lr_total_loss})}
\label{figure_large_learning_rate_forgetting}
\vspace{-0.5cm}
\end{figure}

With Adam optimizer, using the linear scaling rule did not yield any positive results. We found that increasing the learning rate made the training highly unstable and often resulted in the model learning how to play well only to later abruptly forget and score 0 until the end (see figure \ref{figure_large_learning_rate_forgetting}). We settled on using $\eta=0.001$ (the same as in the single-node version), as it was the largest value for which we did not experience large instabilities.

Apart from the learning rate some of the other default optimizer parameters also needed examination. Moving to a synchronous distributed setup requires a re-thinking of how exactly momentum accumulation and learning rate adaptivity are impacted by the batch size.

\subsubsection{Modifying the $\epsilon$ parameter.} 
\begin{figure}[h]
%\centering
\begin{subfigure}[b]{0.5\textwidth}
\input{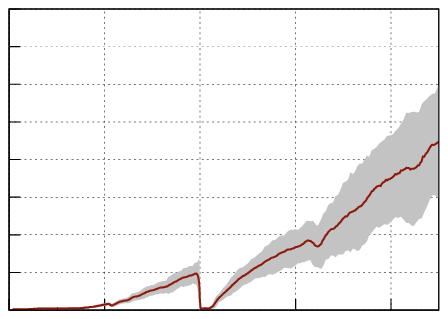}
\caption{First 90 minutes $\hat \epsilon = 10^{-3}$}
\label{epsilon_1e3_90}
\end{subfigure}
~
\begin{subfigure}[b]{0.5\textwidth}
\input{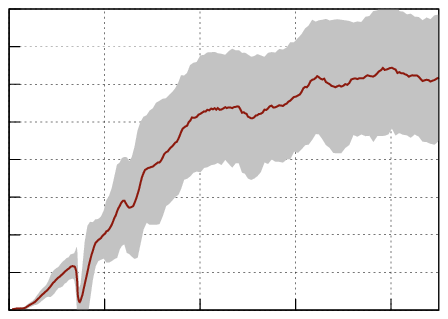}
\caption{First 90 minutes, $\hat \epsilon = 10^{-8}$}
\label{epsilon_1e8_90}
\end{subfigure}

\begin{subfigure}[b]{0.5\textwidth}
\input{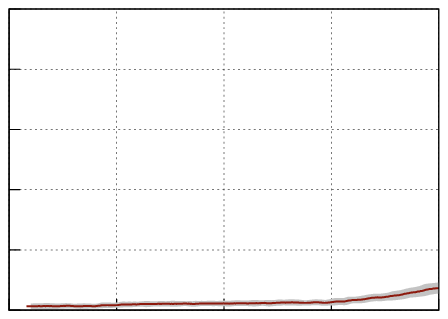}
\caption{First 20 minutes, $\hat \epsilon = 10^{-3}$}
\label{epsilon_1e3_20}
\end{subfigure}
~
\begin{subfigure}[b]{0.5\textwidth}
\input{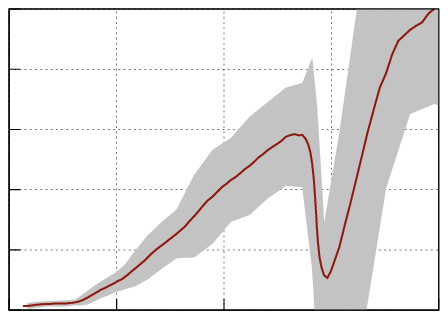}
\caption{First 20 minutes, $\hat \epsilon = 10^{-8}$}
\label{epsilon_1e8_20}
\end{subfigure}
\caption{Two experiments of training agents for Breakout on 64 nodes with different $\epsilon$ parameter values. Figures \ref{epsilon_1e3_90} and \ref{epsilon_1e3_20} show an experiment where $\epsilon = 10^{-3}$ was used. Figures \ref{epsilon_1e8_90} and \ref{epsilon_1e8_20} show training with $\epsilon = 10^{-8}$. The most important difference lies at the beginning of the training. This is visible in the closer views presented in the bottom row. Lower values of $\epsilon$ seem to give a significant speedup at this stage. Note that the vertical axis shows online score.}
\label{epsilon_experiments_figure}
\vspace{-0.5cm}
\end{figure}

 Curiously enough, some implementations can be found that manipulate this variable so that it no longer serves the mere purpose of avoiding numerical instability (see e.g. the implementation of BA3C in \cite{Tensorpack}). Through experiments we found that for some tasks setting the $\epsilon$ parameter of the optimizer to much lower values (e.g. $10^{-8}$ instead of $10^{-3}$) can yield much better training times. A comparison of online scores for two similar experiments with different epsilon values is shown in the figure \ref{epsilon_experiments_figure}. It is important to note that this positive effect when using smaller $\epsilon$ was observable only when using large effective batch sizes (i.e., 512 and more). For smaller effective batch sizes using a lower $\epsilon$ did not produce positive results. 

\begin{wraptable}{R}{5cm}
\centering
\vspace{-0.9cm}
\caption{Number of network parameters when considering different number of neurons in the fully connected layer.}
\begin{tabular}{ |c|c|c| }\hline
\specialcell{hidden\\neurons} & \specialcell{network\\weights} & \specialcell{\% of initial\\setup}\\
\hline
256 & 538 119 & 100\%\\
128 & 332 295 & 61\%\\
64 & 229 383 & 43\%\\
32 & 177 927 & 33\%\\
16 & 152 199 & 28\%\\
\hline
\end{tabular}
%\vspace{-1.5cm}
\end{wraptable}

 This is understandable since a high $\epsilon$ value significantly constrains the ability of the Adam optimizer to automatically adapt the learning rate to the variance of the gradients. We suspected that averaging more data points through the use of synchronous data parallelism reduced the variance of the gradient estimate to the point that the algorithm could be allowed more freedom in automatically adapting the learning rate based on the noise estimations.
 
 Based on these experiments we decided to change the default value of the $\epsilon$ hyperparameter from $10^{-3}$ to $10^{-8}$. This significantly sped up the training for some games (such as Breakout and Boxing). However, we cannot claim this effect is universal, e.g., it made training for Atari Pong slower. The results cited for this game in section \ref{results} were obtained with $\epsilon = 10^{-3}$.
 
\begin{wrapfigure}{R}{6cm}
\vspace{-1.5cm}
\centering
\input{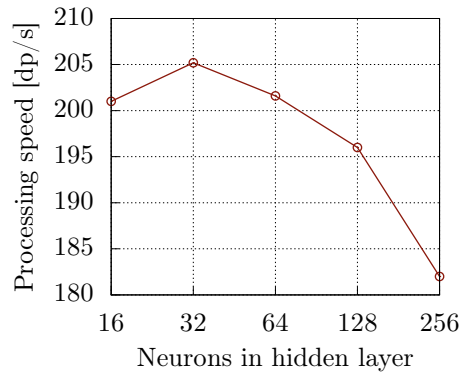}
%\caption{}
\caption{Data points per second for models with different number of hidden neurons in the fully connected layer. Each experiment was repeated 5 times and results were averaged. Experiments were run with 32 workers and 4 parameter servers.}
\label{fig:fc_neurons}
\vspace{-1.2cm}
\end{wrapfigure}

 %In order to confirm this hypothesis we performed a single node experiment with an unusually large batch size to check whether allowing for more adaptability of the learning rate (by setting smaller $\epsilon$) would yield faster training. The results  are shown in the figure \ref{epsilon_single_node_experiment}. Therefore we might conclude that training with large batch sizes (such as when using large scale synchronous data parallelism) calls for smaller values of $\epsilon$.

%Because we were using synchronous training this required us to keep the effective batch size constant across experiments with different numbers of nodes. Otherwise the results would not be directly comparable. 

 \subsubsection{Other hyperparameters.}
 Motivated by the research in \cite{fair_paper} we decided that we should optimize Adam's decay factors $\beta_1,\beta_2$ to the very large batch that we are using. This did not turn out to be an easy task - with the Adam optimizer update policy being quite complicated, choosing $\beta_1$ and $\beta_2$ for the effective batch-size analytically was difficult. The results of our experiments do not support any gains from using different values of these parameters; however we are leaving it to the community to try and find the factors that work best for a distributed setup.

\subsection{Communication overhead} \label{subsection_communication_overhead}

In data parallelized synchronous gradient descent procedures the nodes have to transmit roughly $2n$ multiplicities of the size of the model (where $n$ is the number of workers) during a single training step \cite{MLHPC_scalability_limits}. Our initial model architecture consisted of approximately half a million weights stored as 32-bit floats. Thus, when using 64 workers we needed to send $\approx 263$MB of data during every iteration. % TGREL CHECKED THIS: ((32 / 8) * 538119) * 128. / (2**20) == 262.753

\begin{wrapfigure}{r}{6cm}
\centering
\vspace{-0.7cm}
\input{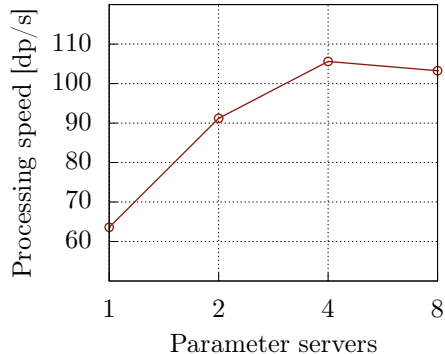}
\caption{Data points per second for different numbers of parameter servers. Each experiment was repeated 5 times and the results were averaged. 32 workers, 332k parameters, local; batch size set to 4. The conclusion is that after some point there's no more gains to be achieved by adding more parameter servers.} 
\label{fig:ps_plot}
\vspace{-0.9cm}
\end{wrapfigure}

Importantly, if all nodes have roughly the same processing speed and synchronous training is being used then all this communication occurs at approximately the same time. This is because in a gradient descent training the communication cannot be easily overlaid with computation to maximally utilize both network bandwidth and compute power (see \cite{MLHPC_scalability_limits} for details). This further reduces the scaling capabilities.

In our experiments we measured the speed of our algorithm by calculating the number of training examples backpropagated through our model every second. In next section we will refer to this speed as data points per second.

\subsubsection{Changing the model.} One way to reduce network communication is to shrink the model. In the initial architecture most weights ($\approx 76\%$) were in a single fully connected layer that follows the last convolution layer (see \cite{ba3c_paper} for the details about the exact neural model used). The relation between number of neurons in this layer and the processing speed is shown in figure \ref{fig:fc_neurons}. Although all of the tested architectures were able to achieve decent results, we decided to use $128$ neurons since this setup was able to learn as fast and stable as the initial architecture, while having only $\approx 61\%$ of its weights. Despite the fact that further reduction of the model size accelerated data processing, smaller networks were taking more time to reach corresponding scores.

\subsubsection{Adding more parameter servers.}

Adding more parameter servers, each storing only fraction of model weights, causes data sent through the network to be distributed into more nodes. This leads to more optimized network usage (see figure \ref{fig:ps_plot} for details).

\section{Results}\label{results}
Synchronous training allowed us to use more workers and avoid instabilities common in the asynchronous paradigm. By reducing model size and adding more parameter servers we could better utilize network communication which led to the possibility of further increase in the number of workers. As a result we were able to train models to reach 300 points in Breakout in $21 \pm 2$ minutes using 64 workers (each consisting of 12 physical cores, i.e. using 768 cores in total).

\subsection{Scaling}
\begin{figure}[h]
\vspace{-0.5cm}
\centering
\begin{minipage}[c]{0.5\textwidth}
\input{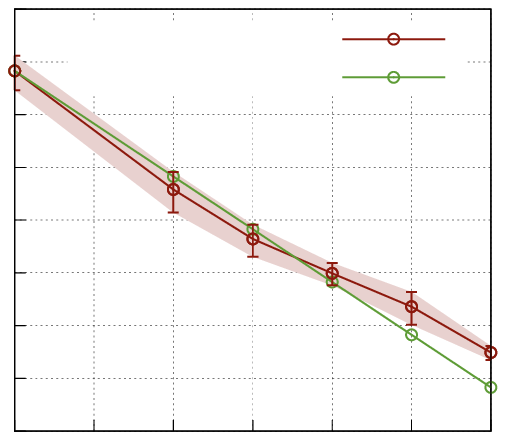}
\end{minipage}
\hfill
\begin{minipage}[c]{0.4\textwidth}
\caption{Red plot shows mean time to reference score of 300 points $\pm$ standard deviation for Breakout. Green plot shows theoretical linear speedup in reference to 1 node experiment. Notice that the real performance of our configuration is consistent with the expected values for a wide number of workers. For 64 workers the communication overheads start to inhibit further scaling. Notice that the mean time to achieve a mean score of 300 in Atari breakout is 21 minutes when using 64 workers.}
\label{fig_scaling}
\end{minipage}
\vspace{-0.5cm}
\end{figure}

We compared times it took to reach a predetermined score in Atari Breakout for different number of workers. The reference was reaching a mean score of 300 points or higher for 50 consecutive games played. This is considered vastly better than a human tester (see \cite{deepmind_nature} for data on detailed human performance for this game). The results are shown in figure \ref{fig_scaling}.

\textbf{Experiment settings:} Each of the 64 workers had 12 CPU cores. We used 4 parameter servers for storing model weights. Model trained had 128 hidden neurons in the fully connected layer described in section 2.3. Every experiment was repeated 10 times and results were averaged. Additionally we have plotted a theoretical linear speedup. This line represents the theoretical time that should be achieved when using $n$ times more computing power in reference to a single node experiment.\\
\textbf{Learning rate:} All experiments used the learning rate of $10^{-3}$.\\
\textbf{Optimizer's hyperparameters:} In all experiments optimizer's hyperparameters were: $\epsilon = 10^{-8}$, $\beta_1 = 0.8$, $\beta_2 = 0.75$.\\
\textbf{Batch size:} In experiments with 32 and 64 workers batch size was set to 32, because smaller batches caused too much network communication overhead. For the rest of experiments the per-worker batch size was set, so that the effective batch size equaled $n\times\texttt{bs}=512$.\\
\textbf{Evaluation:} Every 1000 steps the model played 50 games and the mean score was saved. During the games the model parameters were frozen. By step we mean single global update performed by the chief worker.\\
\textbf{Baseline:} As a baseline we have chosen the single node setup (i.e. using a single 12-core CPU). To be comparable with effective batch sizes on multiple nodes, a relatively large batch size of 512 was chosen. This baseline achieves the solving score in mean time of 14.2 hours.

\subsection{Training times}
In this section we present example learning curves for various Atari games. The plots show mean and max scores from evaluation games. Each game was played on the 64 worker setup.
\begin{figure}[h]
\vspace{-0.5cm}
\begin{subfigure}[b]{0.24\textwidth}
\input{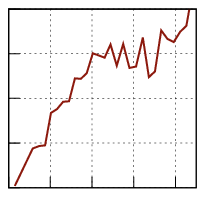}
\caption{Assault}
\end{subfigure}
\begin{subfigure}[b]{0.24\textwidth}
\input{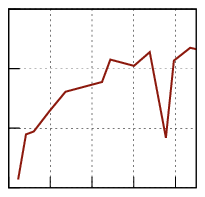}
\caption{BeamRider}
\end{subfigure}
\begin{subfigure}[b]{0.24\textwidth}
\input{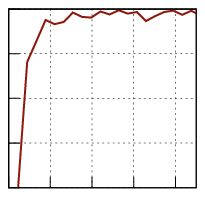}
\caption{Boxing}
\end{subfigure}
\begin{subfigure}[b]{0.24\textwidth}
\input{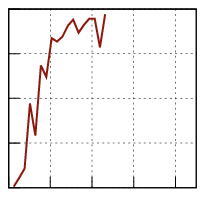}
\caption{Breakout}
\end{subfigure}

\begin{subfigure}[b]{0.24\textwidth}
\input{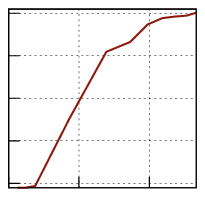}
\caption{Pong}
\end{subfigure}
\begin{subfigure}[b]{0.24\textwidth}
\input{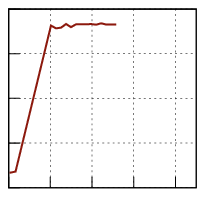}
\caption{Seaquest}
\end{subfigure}
\begin{subfigure}[b]{0.24\textwidth}
\input{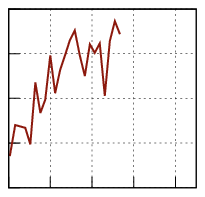}
\caption{Space invaders}
\end{subfigure}
\begin{subfigure}[b]{0.24\textwidth}
\input{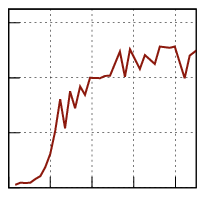}
\caption{Star gunner}
\end{subfigure}
\caption{Score vs time plots for different games in the final setup.}
\end{figure}
\FloatBarrier

\subsection{Comparison with other solutions}
The most notable similar work in optimizing Atari games training for speed is presented in \cite{ga3c_2017}. The results presented there were achieved by a hybrid GPU-CPU algorithm called GA3C which is a flavor of A3C focusing on batching the data points in order to better utilize the massively parallel nature of GPU computations. This is similar to the single node algorithm called BA3C \cite{ba3c_paper} which we used as a starting point for this work.

Comparing the training curves included in \cite{ga3c_2017} for 3 common games tested in both works (Boxing, Breakout, SpaceInvaders) shows that our implementation is much faster and achieves as good or better scores\footnote{It is important to note that the scores achieved by different implementations are not directly comparable and should interpreted cautiously. For future comparisons we'd like to state that the evaluation scores presented by us in this work are always mean scores of 50 consecutive games played by the agent. Unless otherwise stated they're evaluation scores achieved by choosing the action giving the highest future expected reward.}
(see table \ref{ga3c_comparison}). Importantly our experiments used 64 CPU nodes of 12 cores each, while the experiments presented in \cite{ga3c_paper} were all single node. However the results show that using distributed computations on CPU clusters is a viable alternative for using GPUs, even when training convolutional neural networks.

\renewcommand*{\arraystretch}{1.3}
\begin{table}[h]
\centering
\begin{minipage}[c]{0.2\textwidth}
\caption{Algorithm performance in 3 games tested in both papers. Best stable score and time (in hours) to achieve it are given. The data are based on the best reported results found in the training plots in \cite{ga3c_2017,ba3c_paper,deepmind_a3c}.}
\label{ga3c_comparison}
\end{minipage}
\hfill
\begin{minipage}[c]{0.78\textwidth}
\begin{tabular}{ |p{2.3cm}|p{1.8cm}|p{1.5cm}|p{1.5cm}|p{1.7cm}| }
\hline
Game & DDRL A3C& GA3C \cite{ga3c_2017} & BA3C \cite{ba3c_paper} & A3C \cite{deepmind_a3c}\\\hline
BeamRider & 14900 (2.7h) & 3000 (24h)& -- & 15000 (15h) \\
Breakout & 350 (0.5h) & 350 (21h) & 400 (15h) & 500 (11h)\\
Boxing & 98 (0.5h) & 92 (2h) & -- & --\\
Pong & 20 (4h) & 18 (1h) & 17 (24h) & 20 (8h)\\
Seaquest & 1832 (0.5h) & 1706 (24h) & 1840 (24h) & 2300 (24h)\\
SpaceInvaders & 650 (0.5h) & 600 (24h) & 700 (24h) & 1400 (15h)\\\hline
\end{tabular}
\end{minipage}
\end{table}

\section{Conclusions and future work}

We presented a detailed description of our experiments 	 with large scale Distributed Deep Reinforcement Learning (DDRL). Detailed motivation behind all the important design choices was given in sections \ref{subsection_architecture}, \ref{section_sync_async}, \ref{section_optimizers} and \ref{subsection_communication_overhead}. We also provided some empirical information about tuning the Adam optimizer to perform well when using large training batches that arise in synchronous data parallelism. Our key experimental results described in section \ref{results} involve being able to train agents for playing Atari games in minutes rather than hours on clusters of commodity CPUs.

Extending this work to other RL algorithms, most notably those presented in \cite{trpo,ppo,acer} would provide a natural extension to this work. Also developing a framework for distributed RL training that is independent of the algorithm itself would certainly be a valuable contribution.

Given the results reported in \cite{100_epoch_alexnet} testing the Intel\textsuperscript{\textregistered} Xeon Phi$^\mathrm{TM}$ architecture on distributed RL training would also be an interesting experiment.

On a wider scale, further research on adaptive  optimization algorithms, most notably those presented in \cite{adam_paper,adagrad,rmsprop_paper} in the context of training with large batch sizes seems to be necessary to further reduce training times both in supervised and reinforcement learning.  

\section{Acknowledgments}

The work presented in this paper would not have been possible without the computational power of Prometheus supercomputer, provided by the PL-Grid infrastructure.

We would also like to thank the four anonymous reviewers who provided us with valuable insights and suggestions about our work.

This work was supported by the LABEX MILYON (ANR-10-LABX-0070) of Universit\'e de Lyon, within the program "Investissements d'Avenir" (ANR-11-IDEX- 0007) operated by the French National Research Agency (ANR).

\bibliographystyle{splncs03}
\bibliography{bibliography}

\end{document}